\documentclass[journal]{IEEEtran}
\usepackage{graphicx}
\usepackage{amsmath}
\usepackage{amssymb}
\usepackage[numbers, sort]{natbib}
\usepackage{algorithm}

\usepackage{algpascal}
\usepackage{algc}
\usepackage{algcompatible}
\usepackage{algpseudocode}
\usepackage[utf8]{inputenc}
\usepackage[brazilian]{babel}
\usepackage[switch]{lineno}

\begin{document}
%
\title{Accounting for Work Zone Disruptions in Traffic Flow Forecasting}
%
%
%
\author{Yuanjie Lu\thanks{Department of Computer Science, George Mason University, Fairfax, VA, 22030. Email: ylu22@gmu.edu},
\and Amarda Shehu\thanks{Department of Computer Science, George Mason University, Fairfax, VA, 22030. Email: amarda@gmu.edu},
\and David Lattanzi\thanks{Department of Civil Engineering, George Mason University, Fairfax, VA, 22030. Email: dlattanz@gmu.edu}

}
\maketitle


\begin{abstract} 
Traffic speed forecasting is an important task in intelligent transportation system management. The objective of much of the current computational research is to minimize the difference between predicted and actual speeds, but information modalities other than speed priors are largely not taken into account. In particular, though state of the art performance is achieved on speed forecasting with graph neural network methods, these methods do not incorporate information on roadway maintenance work zones and their impacts on predicted traffic flows; yet, the impacts of construction work zones are of significant interest to roadway management agencies, because they translate to impacts on the local economy and public well-being. In this paper, we build over the convolutional graph neural network architecture and present a novel ``Graph Convolutional Network for Roadway Work Zones" model that includes a novel data fusion mechanism and a new heterogeneous graph aggregation methodology to accommodate work zone information in spatio-temporal dependencies among traffic states. The model is evaluated on two data sets that capture traffic flows in the presence of work zones in the Commonwealth of Virginia. Extensive comparative evaluation and ablation studies show that the proposed model can capture complex and nonlinear spatio-temporal relationships across a transportation corridor, outperforming baseline models, particularly when predicting traffic flow during a workzone event.
\end{abstract}

\begin{IEEEkeywords}
Traffic speed prediction, graph neural network, spatio-temporal correlation, hypergraph, work zone, maintenance downtime.
\end{IEEEkeywords}

\section{Introduction}
\label{sec:Intro}

\IEEEPARstart{A}{ccording} to an urban mobility report released in 2019, the economic toll of traffic congestion has increased by nearly 48\% over the past ten years~\cite{schrank_2019_2019,du2017predicting}. Modeling and forecasting traffic flows, including the impacts of maintenance activities (referred to here as work zones) on a transportation corridor, can provide engineers and managers with tools for optimizing the logistics of maintenance while maintaining optimal traffic flow for the traveling public. 

Growing sophistication in deep learning is renewing attention in feature-free intelligent transportation system modeling for  traffic management problems~\cite{Zhang11}. Advances in spatio-temporal modeling and neural network architectures that can handle graph data have lead to many architectures that are increasingly improving performance or extending the prediction horizon on traffic flow~\cite{cui2019traffic, li2017diffusion, yu2017spatio, diao2019dynamic}.

Two main architectures have grown popular in recent years. Convolutional neural networks (CNN) have been employed to extract spatial features of grid-based data and handle high-dimensional spatio-temporal data. Graph convolutional neural networks (GCN) have been shown to be more powerful due to their ability to describe spatial correlations of graph-based data. In this work, we consider the architecture  presented in~\cite{yu2017spatio}, referred to as Spatial-Temporal Graph Convolutional Networks for Traffic Flow Forecasting (STGCN), as representative of the performance of conventional GCN methods. STGCN serves as a baseline model for comparative analysis.

Conventional CNN- and GCN-based models cannot simultaneously characterize the spatio-temporal features and dynamic correlations of traffic data. To address this, a spatial-temporal attention mechanism was added in~\cite{guo2019attention} to learn the dynamic spatial-temporal correlations of traffic data; spatial attention models the complex spatial correlations between different locations, and a temporal attention module captures the dynamic temporal correlations between different times. The model in~\cite{guo2019attention}, referred to as Attention-based Spatio-Temporal Graph Convolutional Networks for Traffic Forecasting (ASTGCN), yields high predictive performance on a variety of related transportation network data sets. For this reason, it serves as the basis for the architecture developed in this work and is a key baseline for comparative analyses.

A key shortcoming of state-of-the-art (SOTA) methods for traffic flow forecasting is the inability to incorporate information modalities other than traffic speed through a segment (graph edge). The result is that SOTA prediction methods, while accurate under normal operating conditions, are not designed to account for work zone impacts within their architectures. Of note is that, because work zone conditions are infrequent relative to normal traffic flow, related traffic disruptions are largely regarded as statistical variance by SOTA methods and are not sufficiently evaluated in prior studies.

In this paper, we build over the SOTA ASTGCN framework and propose a novel GCN-based model for traffic flow forecasting that can additionally account for construction work zones. We refer to the model as GCN-RWZ, which stands for \underline{G}raph \underline{C}onvolutional \underline{N}etwork for \underline{R}oadway \underline{W}ork \underline{Z}ones. 

GCN-RWZ presents several methodological advancements. The model contains novel modules designed to account for workzone disruption conditions. In particular, a novel data fusion mechanism, referred to as a ``speed wave'' is designed to allow for heterogenous inputs into a graph network. A new heterogeneous graph aggregation methodology additionally allows accommodating work zone information in the spatio-temporal dependencies among traffic states. Descriptors of the various traffic network characteristics are included through a flexible “feature map” data format, which enables GCN-RWZ to capture the influence of construction impacts within a corridor of arbitrary scale, making it flexible and generalizable to a variety of corridors and regional conditions.

GCN-RWZ is evaluated on two data sets that capture traffic flows in the presence of work zones. One data set is located in the Tyson’s Corner region of Northern Virginia, and the other in the Richmond region in Virginia's capital city. The two data sets contain traffic speed and corresponding information, and also include granular information regarding the location and duration of work zone events. This information is not available in existing data sources. The Richmond data set is openly available and shared per the Data Availability Statement at the end of the paper.

Extensive comparative evaluation and ablation studies are presented in this work to isolate and evaluate the impacts of workzone disruptions on network-scale traffic speed prediction. Because prior studies did not accommodate work zone information as model inputs, such studies and associated metrics have not previously been established. The evaluation shows that integrating diverse sources of information and characterizing complex and nonlinear spatio-temporal relationships across a transportation corridor leads to improved performance by GCN-RWZ over existing baseline models. More bradly, GCN-RWZ provides transportation managers with the capability to simulate workzones in a roadway corridor and predict the impacts on the traveling public.

The rest of the paper is organized as follows. In Section~\ref{sec:RelatedWork} we provide a focused review of related work on state-of-the-art GCN-based architectures for traffic flow forecasting. The proposed method is then described in detail in Section~\ref{sec:Methods} and evaluated in Section~\ref{sec:Experiments}. The paper concludes with a summary of future work in Section~\ref{sec:Conclusions}. 
\section{Related Work}
\label{sec:RelatedWork}
Shallow machine learning models have been widely used to predict city-scale traffic flow, for instance ARIMA models~\cite{tong2008highway}, support vector regression~\cite{wu2004travel}, hybrid ensemble models including ANNs and bagging~\cite{moretti2015urban}, and spatial auto-regressive (SAR) models \cite{kelejian1999generalized}. Due to their abilities to incorporate disparate data types through expanded dimensionality, and to handle nonlinear data associations, these models can capture the spatial and temporal correlation in traffic data. However, their reliance on expert-crafted features is a key limitation that hampers their performance, generalizability, and adoption.

In response, researchers have turned to deep learning models that learn directly from data. Early work in traffic prediction has focused on deep belief networks~\cite{huang2014deep}, recurrent neural network (RNN)~\cite{tian2015predicting}, and long short-term memory (LSTM)~\cite{ma2015long} models that can model temporal correlations. More recently, researchers have used multi-model patterns to consider both temporal and spatial dependencies. For instance, work in~\cite{ma2015large} utilizes a deep Restricted Boltzmann Machine within an RNN architecture to capture features of traffic congestion. Work in~\cite{cui2018deep} proposes a deep bidirectional and unidirectional LSTM framework to measure backward dependencies. Recently, many researchers have been inspired by the capability of CNN-based frameworks (in the computer vision domain) to extract structured features and so have utilized CNN-based model to capture spatial correlations between traffic sensors. Work~\cite{ma2017learning} proposes a CNN-based method that models traffic as a set of large-scale, network-wide images. Subsequently, work in~\cite{jo2018image} develops a CNN to convert traffic states into an enhanced physical map. While these methods have improved prediction accuracy, they do not easily capture spatial relationships across a transportation corridor.

Graph neural networks (GNNs), a recent advancement in deep learning, can capture spatial correlations and are now popular in natural language processing (NLP), image, and speech recognition~\cite{vaswani2017attention}. Their utilization in transportation engineering is emerging~\cite{wu2019graph, pan2019urban, ruiz2020gated, zheng2020gman, keneshloo2019deep, velivckovic2017graph}. Since a graph is composed of a complex of atomic information fragments and uses its structural links to represent the relationship between entities, it can be applied to the topology of a road-level traffic network via the concept of a graph and so capture both spatial and temporal correlations~\cite{bacciu2020gentle}. Some GNN-based transportation research utilizes graph network embedding~\cite{perozzi2014deepwalk, kang2019learning, zheng2020gman} and recurrent graph neural networks (RecGNNs)~\cite{wu2020comprehensive, ruiz2020gated}. In addition to being computationally costly, these methods only transmit the information of each node and update the state of its own node, which cannot capture spatial relationships in a traffic network. 
To address this, current state-of-the-art methods use a particular network variant, the Graph Convolutional Network (GCN). Instead of iterating over states and propagating information from a sequence of nodes, GCNs attempt to support a graph with a fixed structure and build convolutional layers to extract the essential features. Such a model, pioneered in~\cite{zhao2019t}, predicts traffic speeds by combining GCN and the Gated Recurrent Unit (GRU) model; the GCN is used to learn topological structures for capturing the spatial correlations, and the GRU is used to learn variations of each tensor for capturing the temporal dependencies. In~\cite{zhou2020reinforced}, a new policy gradient is also proposed for updating the model parameters while alleviating bias.

Learning on graphs is challenging, especially with respect to heterogeneous data. The structure of graphs can capture the dependencies between entities in planar vectors, and by connecting the corresponding graph structures, data information from different applications can be naturally merged. Recently, designing methods for automatic feature extraction has become an important field of graph research~\cite{oneto2022towards}. In order to integrate many blocks in a graphical traffic network, some studies propose models to integrate existing traffic components, thus improving the accuracy of the models. For example, work in~\cite{guo2019attention} introduces three independent temporal components to capture recent, daily-periodic, and weekly-periodic dependencies, and then fuses them to generate the traffic flow. Work in~\cite{du2020traffic} provides a dynamic transition convolution unit to capture the evolution of the demand dynamics so that the model is able to incorporate the spatio-temporal states of the traffic demands with different environmental factors. Work in~\cite{zhu2022kst} provides a knowledge fusion cell to combine the traffic features as the input of the graph neural network, so that the model can capture the impact of external factors on the traffic environment. 

Related work in the rail transit domain \cite{luo2022evaluating} leverages a graph neural network model that encodes the acoustic data collected by a microphone array and incorporates the physical background of sound in the Doppler effect and acoustic attenuation to evaluate railway noise sources. In a similar rail corridor study, \cite{li2022graph}, the authors provide a generator network to extract urban rail transit information and a spatio-temporal discriminator to identify passenger flow predictions from the generator network so that the graph network can be optimized by adversarial loss to enhance the robustness of the model in urban rail transit.

GCN-based methods have shown their capability for traffic forecasting across many domains, and modern methods are able to consider sophisticated and multi-faceted representations of traffic flow. However, these studies primarily use traffic speed, or characterizations of traffic speed, for flow prediction. No prior efforts consider how construction workzone information, which is both temporally and spatially sparse, can be incorporated within a GCN framework. Furthermore, no prior studies have explicitly evaluated network performance for the case of construction workzone disruptions. Because existing data sets do not include information about active workzones, the impact of these events manifests as unquantified statistical variance in traffic flow prediction. 

This study addresses these limits through a novel data fusion mechanism and a new heterogeneous graph aggregation methodology to accommodate work zone information. Furthermore, this study provides new comparative analysis methods designed to isolate and characterize GCN model performance under short-term disruptions, a process that has not been employed in prior studies. Achieving this required the creation of new data sets that includes workzone information in tandem with historic traffic flow predictions. One of these data sets from the Richmond, Virginia, region is shared with the research community as an additional contribution. 
\section{GCN-RWZ: Architecture and Methodology}
\label{sec:Methods}

\begin{figure*}[htbp]
\centering
\includegraphics[width=\textwidth]{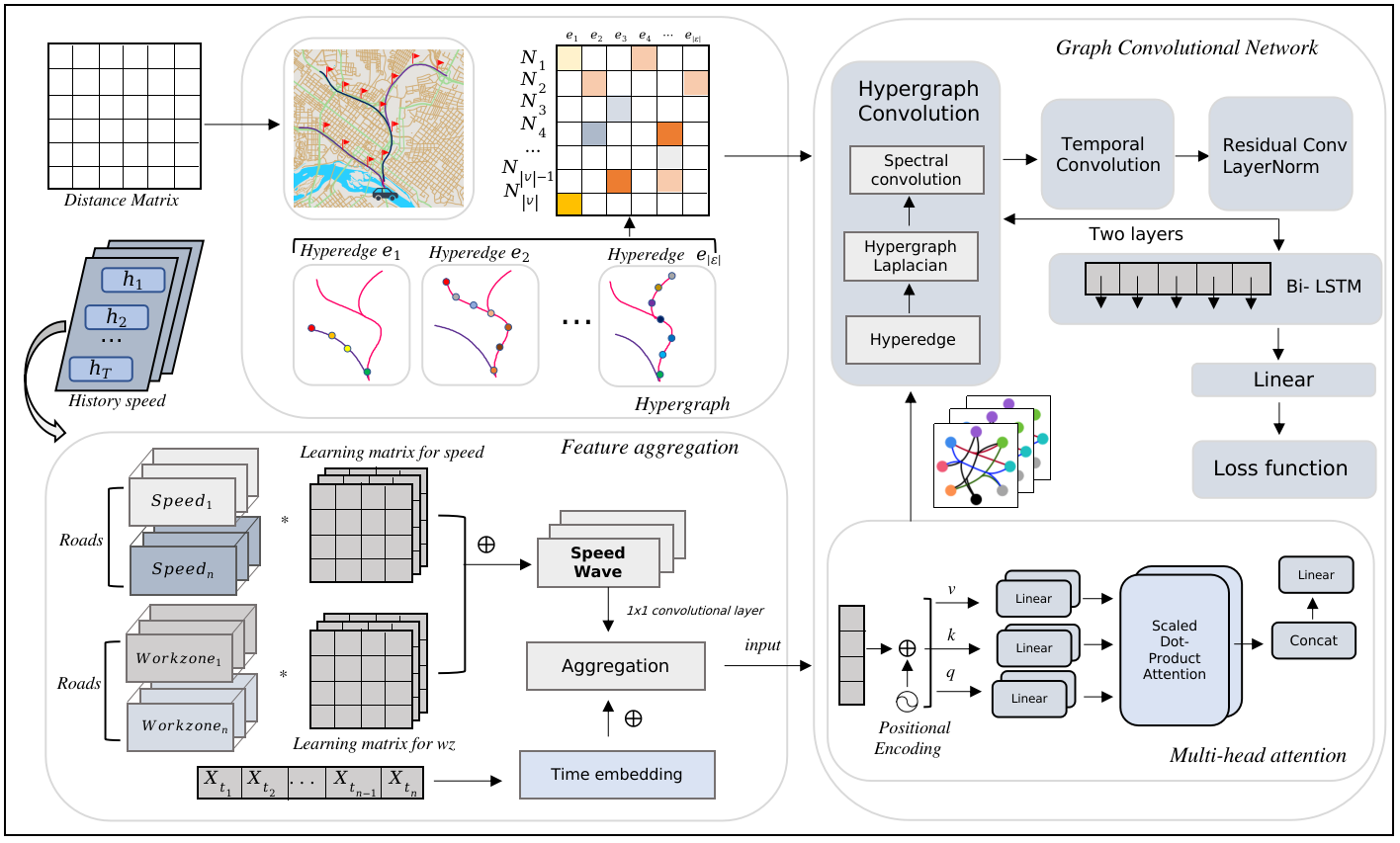}\\

\caption{Overall framework of the developed speed prediction methodology and of the proposed GCN-RWZ model.}
\label{Framework}
\end{figure*}

Our model GCN-RWZ is implemented based on the baseline design of ASTGCN~\cite{guo2019attention}, which merges three different time series segments to capture dynamic spatio-temporal correlations. The difference between our model and theirs is that we focus on fusing data features and then feeding the information fusion into a hypergraph convolution network to learn partial spatial and temporal dependencies. As summarized conceptually in Figure~\ref{Framework}, our model consists of a feature aggregation algorithm and a hypergraph convolution network with a multi-head spatio-temporal attention mechanism. The left panel in the figure shows how the model ingests a variety of different feature maps associated with a range of network descriptors; time series data representing the traffic flow in a corridor are fused with corresponding information on construction work zones. The right panel of the figure shows that the data generated by feature fusion is input into a graph convolutional network that includes two layers of attention mechanisms, hypergraph convolution operations and temporal operations. The information transfer between spatial and temporal dependencies is then fed into a bidirectional recurrent neural network to process the forward and backward dependencies of each node in the time series. 

\subsection{Road Network as Graph}
The traffic network is represented as a hypergraph $G = (\nu, \varepsilon, W)$, where the vertex set $\nu$ includes $N$ road segments and the edge set $\varepsilon$ includes a subset of neighbor roads $\{N_1, N_2, \dots, N_n\}$ defined by geographic distance. Each hyperedge $\varepsilon$ is assigned with a diagonal matrix of edge weights, written as $W \in \mathbb{R}^{|\varepsilon| \times |\varepsilon|}$. The hypergraph of traffic prediction $G$ can be defined as an incidence matrix $H \in \mathbb{R}^{|\nu| \times |\varepsilon|}$, with entries defined by the function $h(v,e) = 1$ only if $v \in e$, otherwise $h(v,e) = 0$. Next, the degrees of each vertex $v \in \nu$ are written as a function: $d(v) = \sum_{e \in \varepsilon}{w(e)h(v,e)}$ and the degrees of each edge $e \in \varepsilon$ are defined as: $\delta(e) = \sum_{v \in \nu}{h(v,e)}$. Propagated by upstream and downstream roads, we also define a road-based adjacency matrix $A = (A_{i,i}, \cdots, A_{n,n}) \in \mathbb{R}^{N \times N}$ to record the connectedness of nodes in $G$. In addition, we define a distance-based matrix $D$ to record the geographic distance between each node and other downstream nodes.

\subsection{Heterogeneous Graph Aggregation}
We leverage the concept of a heterogeneous graph to build a feature map and perform data fusion. While we consider the problem of heterogeneous data fusion in the context of construction work zones, our approach could be extended to other sources of information such as traffic speed, the number of lane closures, historical information, or weather, among others.

A heterogeneous graph is an information network that can incorporate multiple types of objects or multiple types of links~\cite{wang2019heterogeneous}. Based on the graph design, we provide a novel methodological component that leverages the notion of a feature map, $X$, to account for generalized information. The feature map corresponding with the defined graph structure is written as $X = (X_{i,j}, \cdots, X_{n, k}) \in \mathbb{R}^{N \times T}$, where $N$ is the number of road segments as above, $T$ is the length of the whole time series and $X_{i,j}$ represents the value of a feature $X$ at the $i^{th}$ node at the $j^{th}$ time step. All feature maps $X$ have same dimensions in the graph. 
\subsubsection{Speed feature map $X^{S}$}
Speed modeling is based on time series forecasting that predicts values over a period of time based on historical data. In traffic forecasting, the speed $X^S_{i,j}$ of traffic at vertex $i$ during time $j$ is assumed to be impacted by the speeds $X^S_{t_1, \dots, t_{j-1}}$ for prior time steps $t_1$ to $t_{j-1}$ and the speeds of all road segments $v \in V$ at time step $t_j$. Two hyper-parameters $H$ and $P$ indicate the length of the time series for training and prediction, respectively, in terms of the number of time steps.
\subsubsection{Average history speed feature map $X^{AS}$}
The average historical speed feature map has two advantages: it is able to effectively smooth time series data in the presence of missing values, and it is able to quantify abnormal traffic events. The history speed feature map is defined as $X^{AS}_{i,j} = \frac{1}{n}\sum_{t=1}^{n}X^S_{t,j}$, where $t$ represents the same time interval in a day and in a same week and $X^S_{t,j} \neq 0$. For example, the historical speed of 8 o'clock on Tuesday is the average speed for all 8 o'clock on Tuesday in a year.

\subsubsection{Diff feature Map $X^{D}$} 
The diff feature map is mainly used to measure the difference between the speed $X^S_{i,j}$ of traffic at road $i$ during time $j$ and the corresponding average history speed $X^H_{i,j}$, which is written as $X^D_{i,j}$ = $X^S_{i,j}$ - average history speed $X^{AS}_{i,j}$. The diff feature map is designed to identify if the traffic environment is affected by a disruption. If $X^S_{i,j}$ is greater than $X^{AS}_{i,j}$, we heuristically believe that, regardless of the presence of anomalies, the factor has no effect on traffic. The advantage of this is that some data that have no effect on the actual traffic environment are excluded so that only those abnormal points that really have an impact on traffic are considered by the model. There is a threshold $\delta$ to determine $X^D_{i,j}$. If $X^D_{i,j} > \delta$, we define a normal traffic environment at road $i$ during time $j$. Otherwise, it is defined as abnormal traffic environment.

\subsubsection{Time matrix}
The traffic speed curve changes very obviously with time in a day and in a week, such as peak time with off peak time, weekday with weekend. Therefore, according to the time interval of speed observations, we assign the time of the week into corresponding bins and then use an approximate embedding mechanism to represent continuous values with learning weights. In this way, the model can estimate the effect of traffic speed changes over time.

\subsubsection{Construction feature map $X^{C}$} A construction work zone map, $X^C_{i,j}$, is used to record whether the road $i$ has a work zone event at time $j$.  A purely binary  feature creates numerical problems due to the sparsity of the resulting map, and it ignores the impacts of a work zone on neighboring road segments. An efficient way to solve this problem is to use radial basis functions to measure the relationship between roads. If there is road construction at road $i$ during time $t$, GCN-RWZ defines the scoped area in advance and then evaluates how the event impacts the area. Not all work zone events impact traffic, especially under low traffic volume. To quantify this, we combine a binary diff feature map and the construction feature map, to ignore those $X^C_{i,j}$ whose diff $X^D_{i,j}$ is larger than the threshold $\delta$. 

\subsubsection{Feature aggregation} To measure the weight ratio of each feature map, we define a speed wave function, written as $\hat{X_s} = W_s \odot X^s + W_c \odot X^c$, where $\hat{X_s}$ denotes speed wave, $\odot$ is the Hadamard product, and $W_s$ and $W_c$ are learning parameters reflecting the influence degrees of features on traffic states. We use a 1x1 convolutional layer to increase its dimensionality, and the Hadamard product to fuse the time feature map. The result is that the GCN-RWZ model is able to detect if there are construction work zone events during model training and if so, it evaluates the impact of those events on the surrounding road segments. 

\subsection{Hypergraph convolution network} 
\subsubsection{Multi-head Spatial-Temporal Attention Mechanism} GCN-RWZ uses a multi-head attention mechanism to learn  spatio-temporal dependencies. The core concept is to allow a decoder to exploit the most relevant parts of the input sequence in a flexible way and assign the highest weight to the most relevant vectors. In graph structure, the spatial-temporal attention mechanism allows the neural network to pay more attention to more valuable information. The input, adjusted by the attention mechanism, is then fed into the spatial-temporal convolution operations. Graph convolution operates over the spatial dimension to capture spatial dependencies and temporal convolution operates over the temporal dimension to capture temporal dependencies.

\subsubsection{Spatial hypergraph convolution network}
To learn the topological relationships in a traffic network, graph convolutional operations are performed on the input feature map from the training data. A Laplacian matrix of the graph is constructed to derive the Laplacian operator and then perform eigendecomposition by Fourier transform. This decomposition represents the signal over the graph $G$ at time $t$ as $x = \bf{x}_t^f \in \mathcal{R}^N$. The graph Fourier transform of the signal is then $\hat{x} = U^T x$. Since $U$ is an orthogonal matrix, the corresponding inverse Fourier transform is $x = U\hat{x}$. The signal $x$ on the graph $G$ is filtered by a kernel $g \in R^N$, and the graph convolutions are defined as:
\begin{equation}
x \ast g = f^{-1}(f(x)\odot f(g)) = U(U^{\mathsf{T}} x \odot U^{\mathsf{T}} g),
\end{equation}
where $\ast$ is graph convolution operation and $\odot$ is the Hadamard product. If we define $U^{\mathsf{T}} g$ as $g_{\theta}$, which is a learnable convolution kernel, the graph convolution is written as: $(x \ast g)_G = U g_{\theta} U^{\mathsf{T}} x
$. 

The computational cost of this operation is high, because each sample needs feature decomposition, and each forward propagation needs to calculate the product of $U$, $g_{\theta}$, and $U^{\mathsf{T}}$. Inspired by work in~\cite{defferrard2016convolutional, hammond2011wavelets}, the $g_{\theta}$ can be expanded by Chebyshev polynomials, defined as: 
\begin{equation}
g_{\theta}(\Lambda)\approx \sum_{k=0}^{K} \theta_k T_K (\widehat{\Lambda}),
\end{equation}
where $\widehat{\Lambda} = \frac{2\Lambda}{\lambda_{max}} - I_N$, $\lambda_{max}$ is the spectral radius, $\theta$ is the vector of Chebyshev coefficient, $T_K$ is defined as $T_k(x) = 2xT_{k-1} - T_{k-2}(x)$, where $T_0(x) = 1$ and $T_1(x) = x$. 
Thus, the graph convolution operation is denoted as:
\begin{equation}
(x \ast g)_G = \sum_{k=0}^{K} \theta_k T_K (\widehat{L})x,
\end{equation} 
where $\widehat{L} = \frac{2L}{\lambda_{max}} - I_N = U\widehat{\Lambda}U^{\mathsf{T}}$. Due to the expansive computation of Laplacian Eigenvectors and higher computation complexity, we use a layer-wise linear model as in~\cite{kipf2016semi}, where $K$ = 1 and $\lambda_{max}$ = 2 to limit the order of convolution operations and improve the scale adaptability of neural networks. The convolution operation can be further simplified as:
\begin{equation}
(x \ast g)_G = \theta_0 x - \theta_1 D_v^{-1/2}HWD_{e}^{-1}H^{\intercal}D^{-1/2}_vx
\end{equation}
Subsequently, we set $\theta_0 = \frac{1}{2} \theta D_v^{-1/2}HD_e^{-1}H^{\intercal}D^{-1/2}_v$ and $\theta_1 = -\frac{1}{2}\theta$ to avoid overfitting. The convolution operation can be written as: 
\begin{align}
(x \ast g)_G &= \frac{1}{2}\theta D_v^{-1/2}H(W+I)D_e^{-1}H^{\intercal}D^{-1/2}_vx \\
& = \theta D_v^{-1/2}HWD_e^{-1}H^{\intercal}D^{-1/2}_vx,
\end{align}
where $W$ is an identity matrix and ($W+I$) is regarded as weights of hyperedges.

\subsubsection{Temporal convolutional}
After spatial convolution is carried out, temporal dependencies are computed through a standard convolutional operation per \cite{zhao2019t}. The output after the spatial-temporal convolution is written as: 
\vspace*{-2mm}
\begin{equation}
\Bar{X}^{(l+1)}_H = \sigma(\Phi \ast (\sigma((x \ast g)_G X^{(l)}_H))) \in R^{C \times N \times T},
\end{equation} 
where $\Phi$ is a parameter of the temporal convolution kernel, the first $\ast$ is a standard convolution operation, $(x \ast g)_G$ represents the graph convolution operation, and $\sigma$ is the ReLU activation function. 

\subsubsection{Residual work and bi-RNN}
After all convolutional operations are completed, the resulting output $\Bar{X}^{(l+1)}_H$ of the convolution is fed back through an additional layer of spatial and temporal convolutions. A 1x1 convolutional layer is then used to reduce the output channel to a single dimension. Subsequently, a bidirectional recurrent neural network is used to learn the dynamic behavior in the time sequence for each node. A final linearization function ensures that the output has the same dimension as the prediction.

\subsubsection{Overall process of training GCN-RWZ}
The main procedures of training GCN-RWZ are summarized in Algorithm \ref{alg:framework}. The algorithm generates the required feature maps and then aggregates them as a speed wave input. This speed wave is then combined with the time embedding and input into a two-layer hypergraph neural networks that returns a prediction of future traffic flow throughout a transportation network. The model is trained by backpropagation in GNN.

\begin{algorithm}
\caption{Pseudo-code of GCN-RWZ framework}\label{alg:framework}
\textbf{Input} Speed feature map $X^S \in \mathbb{R}^{N \times T}$, binary construction work feature map $X^{BC} \in \mathbb{R}^{N \times T}$, diff feature map $X^{D} \in \mathbb{R}^{N \times T}$, time feature map $X^{T} \in \mathbb{R}^{N \times T}$, distance matrix $D \in \mathbb{R}^{N \times N}$, adjacency matrix $A \in \mathbb{R}^{N \times N}$, time embedding $E$.
    \begin{algorithmic}[1]
    \State Historical feature map $X^H$ $\gets$ $X^S$ and $X^{D}$
    \State Fix missing value in $X^S$ and set mask of $X^S$
    \State Representative work zone feature map $X^C$ $\gets$ $X^{BC}$ and $X^D$
    \State Training data $X$ $\gets$ $X^S$, $X^C$, $X^T$, mask of $X^S$ and $X^{D}$
    \For{\textit{each batch of $X$}}
        \State $\hat{X_s} = W_s \odot X^s + W_c \odot X^c $
        \State Compute spatial multi-head attention and temporal multi-head attention based on $\hat{X_s}$ and $E$ 
        \State Compute the spatial-temporal convolution operation $\sigma(\Phi \ast (\sigma((x \ast g)_G X^{(l)}_H)))$ 
        \State Compute the residual convolution after operations
        \State Repeat as line 7 into the second layer of graph neural network
        \State Compute the residual work in $1 \times 1$ convolutional layer
        \State Compute the output $\Bar{X}^{(l+1)}_H$ and sent to bidirectional recurrent neural network
    \EndFor 
    \end{algorithmic}
\end{algorithm}
\section{Experiments}
\label{sec:Experiments}

\begin{figure*}[htbp]
  \centering
  \begin{tabular}{cc}  
  \includegraphics[width=0.93\columnwidth]{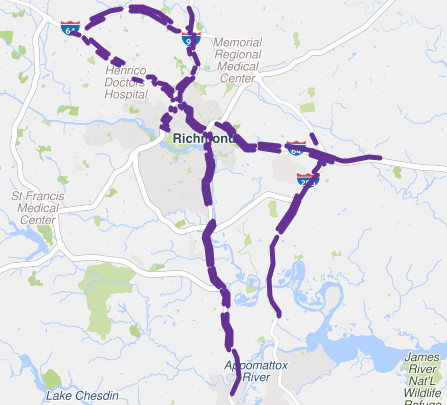}
  \includegraphics[width=0.97\columnwidth]{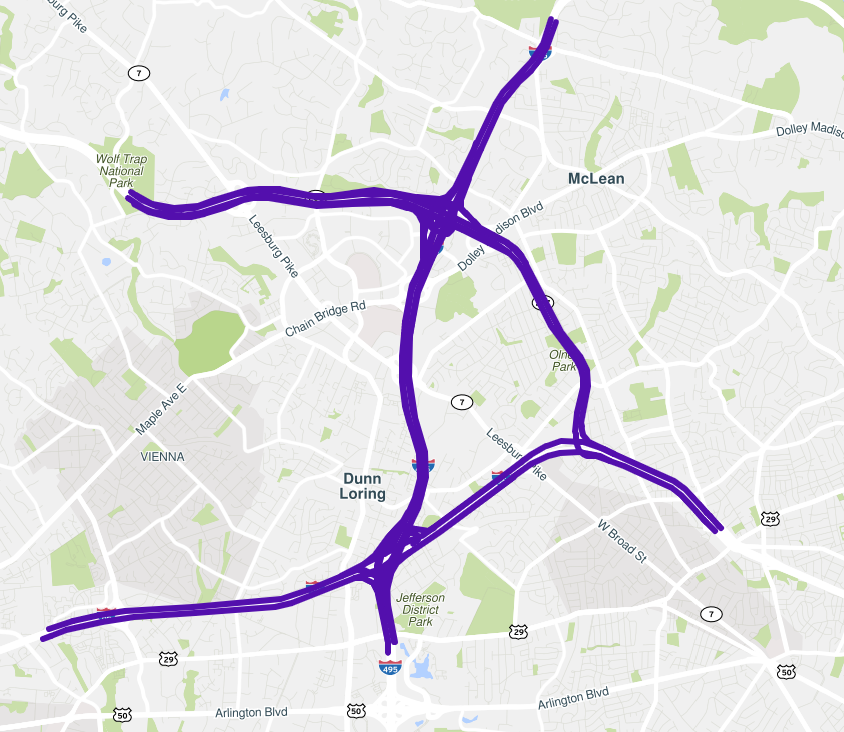} &
  \\[-3mm]
  \end{tabular}
   \caption{Left: The Richmond data set is derived from an in-roadway sensor network in Richmond, Virginia. Right: The Tyson's data set is derived from RITIS data in the Tyson's Corner region of Northern Virginia.}
  \label{data sets}
\end{figure*}

\subsection{Data Sets}
As mentioned previously, there are no openly accessible data sets that contain granular workzone information synchronized with historical traffic flow speed through an associated network. To address this, two different data sets representing urban regions in the state of Virginia were created. One data set is centered on the Richmond metropolitan area. The other is from Tyson's Corner, in the Northern Virginia region. These data sets are shown in Figure \ref{data sets}. Table~\ref{Statitics} provides basic statistics about each data set. The Richmond data set is derived from in-roadway speed sensor data combined with workzone data provided by the Virginia Department of Transportation (VDOT). The data set includes 95 road segments, with data collected from 01/01/2019 to 12/31/2019 at a 15-minute sampling interval. During this time there 1896 work zone events across the network. The Tyson's data set is aggregated from the Regional Integrated Transportation Information System (RITIS) for 131 road segments. The time interval for this data set is also 01/01/2019 to 12/31/2019, but the sample rate is 5-minute intervals. There were 1537 construction work zones within this network over the specified time interval. In the Richmond data set, any road segment without sufficient real time data will record a speed of 0 miles per hour (MPH). However in the Tyson's data set, these events use the historic average speed for a given segment during that time period. Both data sets are cleaned to remove anomalous events, such as traffic accidents concurrent with workzone activity, in order to better isolate the key variable: work zone impact on network traffic flow. Min-Max normalization of the data controls for data imbalance.

In order to properly evaluate the impact of work zones on traffic speed predictions, the data is segmented into normal operating conditions where no work zones are present and disrupted operating conditions when a work zone is present. This segmentation was only used for model evaluation and the complete data record is employed for model training. 

\begin{table}[!htbp]
  \renewcommand\arraystretch{1.3}
   \small
    \centering
  \setlength\tabcolsep{10pt}
\caption{Statistics of data sets}
\vspace{1mm}
\label{Statitics}
 \begin{tabular}{p{3.5em}|p{7.5em}|p{6.5em}|p{2em}} 
 \hline
data sets & Work Zone Events & Sample Rate & Nodes  \\ [0.1ex] 
 \hline
Richmond & 1896 & 15 mins & 95\\ 
 \hline
Tyson's  & 1537 & 5 mins & 135\\
 \hline
\end{tabular}
\end{table}

\subsection{Performance Metrics}
Traffic speed prediction performance was quantified using three metrics: Root Mean Squared Error (RMSE), Mean Absolute Error (MAE), and Mean Absolute Percentage Error (MAPE). The units for RMSE and MAE, which measure the error between predicted and ground-truth speed in a network segment, is miles per hour. In contrast, MAPE considers not only the error between the predicted and true speed, but also the ratio of the error to the true value. 

\subsection{Evaluation Setting}
We compare GCN-RWZ to 4 baseline approaches:  STGCN~\cite{yu2017spatio}, GraphWaveNet~\cite{wu2019graph}, ASTGCN~\cite{guo2019attention} and DGCRN~\cite{li2021dynamic}. As discussed in Section~\ref{sec:Intro}, STGCN represents the baseline spatio-temporal GCN platform that many other methods, such as ASTGCN, extend. GraphWaveNet uses an adaptive dependency matrix and node embedding to capture the hidden spatial dependency in the data and then feeds the information to a dilated graph convolution; ASTGCN uses Chebyshev polynomials and a standard CNN to learn traffic speeds at three different times components, and then integrates traffic information to automatically assign weight ratios to each temporal component. DGCRN designs hyper networks to leverage and extract dynamic characteristics node attributes, then filter the node embeddings to generate a dynamic graph. It is emphasized that while STGCN, GraphWaveNet, ASTGCN, and DGCRN are representative SOTA methods, none of them can account for construction work zones. As described in Section~\ref{sec:Methods}, GCN-RWZ is based on the established ASTGCN architecture, with several key extensions designed to augment performance during work zone events.

\subsection{Performance Comparison}
Table~\ref{tab:Richmond_clean} and Table~\ref{tab:Richmond_wz} relate MAE, (Mean) RMSE, and MAPE for the Richmond region for forecasts of $45$-, $90$-, and $180$-minutes. This corresponds to prediction intervals of 3, 6, and 12 time-steps given the 15-minute sampling rate. Table~\ref{tab:Richmond_clean} shows relative predictive performance on the network under traffic normal conditions.  The results in Table~\ref{tab:Richmond_wz} shows predictive performance during work zone disruptions. 

Table~\ref{tab:Tyson1} and Table~\ref{tab:Tyson2} convey MAE, (Mean) RMSE, and MAPE for the Tyson's Corner data set for forecasts of $15$-, $30$-, and $60$-minutes. This corresponds to prediction intervals of 3, 6, and 12 time-steps given the 5-minute sampling rate in this data set. Table~\ref{tab:Tyson1} shows relative predictive performance on the network under normal conditions. Table~\ref{tab:Tyson2} shows predictive performance during work zone disruptions. 

\begin{figure*}[htbp]
  \centering
  \begin{tabular}{cccc}
  \includegraphics[width=\columnwidth]{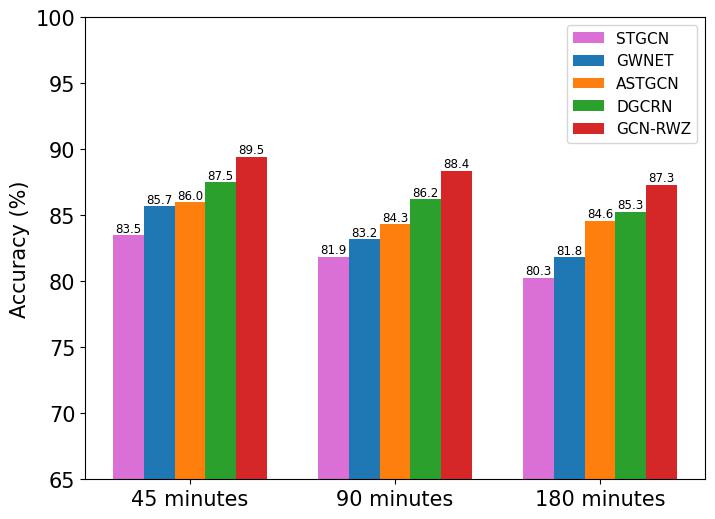} &
  \includegraphics[width=\columnwidth]{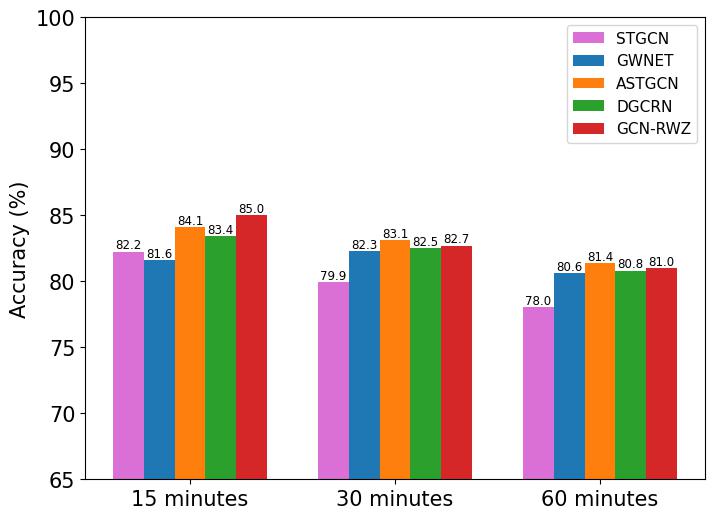}\\[-3mm]
  \end{tabular}
   \caption{Predictive accuracy during workzone disruptions where the ground truth change in speed relative to historical average is greater than $\pm 5$ MPH. Forecast shown for 3, 6, and 12 time-step predictions. Left: Richmond data set (15-minute sample interval). Right: Tyson's data set (5-minute sample interval)}.
  \label{fig:accuracy}
\end{figure*}

As an additional performance metric, we isolate only those instances where predicted speed deviates from ground truth by $\pm 5$ MPH during work zone events. This allows a clearer demonstration of how the models behave under notable disruptions and eliminates the effect of work zones that did not noticeably impede traffic flow (Figure~\ref{fig:accuracy}).

\begin{table*}[h]
    \renewcommand\arraystretch{1.1}
    \small
  \centering
  \setlength\tabcolsep{3.5pt}
  \caption{Relative model performance on Richmond data set during normal operating conditions; best value per metric is highlighted in boldface font}
  \label{tab:Richmond_clean}
  \begin{tabular}{|p{3.1em}|p{6.2em}|p{2.5em}p{2.5em}p{5em}|p{2.5em}p{2.5em}p{5em}|p{2.5em}p{2.5em}p{5em}|}
    \hline
    data set & Model 
    & \multicolumn{3}{c|}{45-min forecast}
    & \multicolumn{3}{c|}{90-min forecast}
    & \multicolumn{3}{c|}{180-min forecast} \\
    \cline{3-11}
      & & MAE& RMSE & MAPE  (\%) & MAE& RMSE & MAPE (\%) & MAE& RMSE & MAPE (\%)
    \\ 
    \hline
&STGCN & 1.60 & 3.69 & 4.21 & 1.73 & 3.99 & 4.93 & 1.92 & 4.35 & 5.52  \\

 & GraphWaveNet & 1.39 & 3.35 & 3.91 & 1.56 & 3.77 & 4.94 &1.75 & 4.19 & 5.42 \\
RICH &ASTGCN  & 1.41 & 3.33 & 3.88 & 1.52 & 3.55 & 4.46 & 1.54 & 3.63 & 4.26   \\
MOND &DGCRN  & 1.25  & 3.05  & 3.29  & 1.35   &  3.38  &  3.76  &  1.43   &  3.63  & 4.04  \\
&GCN-RWZ  & \textbf{1.09} & \textbf{2.68} & \textbf{2.27}& \textbf{1.16} & \textbf{2.92} & \textbf{2.46} & \textbf{1.29} & \textbf{3.14} & \textbf{2.74}\\
    \hline
\end{tabular}
\end{table*}

\begin{table*}[h]
    \renewcommand\arraystretch{1.1}
    \small
  \centering
  \setlength\tabcolsep{3.5pt}
  \caption{Relative model performance on Richmond data set during work zone disruptions; best value per metric is highlighted in boldface font}
  \label{tab:Richmond_wz}
  \begin{tabular}{|p{3.1em}|p{6.2em}|p{2.5em}p{2.5em}p{5em}|p{2.5em}p{2.5em}p{5em}|p{2.5em}p{2.5em}p{5em}|}
    \hline
    data set & Model 
    & \multicolumn{3}{c|}{45-min forecast}
    & \multicolumn{3}{c|}{90-min forecast}
    & \multicolumn{3}{c|}{180-min forecast} \\
    \cline{3-11}
      & & MAE& RMSE & MAPE  (\%) & MAE& RMSE & MAPE (\%) & MAE& RMSE & MAPE (\%)
    \\ 
    \hline
&STGCN & 1.73 & 3.84 & 4.90 & 1.90 & 4.19 & 5.40 & 2.12 & 4.61 & 6.17  \\

& GraphWaveNet & 1.52 & 3.91 & 4.40 & 1.73 & 4.05 & 5.07 & 1.92 & 4.48 & 5.86 \\
RICH &ASTGCN  & 1.53 & 3.51 & 4.36 & 1.67 & 3.82 & 4.86 & 1.68 & 3.83 & 5.04 \\
MOND &DGCRN  & 1.35  & 3.26  & 4.03  &  1.45  &  3.59  &  4.48  & 1.53   &  3.82  &  4.87  \\
&GCN-RWZ  & \textbf{1.16} & \textbf{2.88} & \textbf{2.52} & \textbf{1.25} & \textbf{3.13} & \textbf{2.78} & \textbf{1.36} & \textbf{3.34} & \textbf{3.06}\\
    \hline
\end{tabular}
\end{table*}

\begin{table*}[h]
    \renewcommand\arraystretch{1.1}
    \small
  \centering
  \setlength\tabcolsep{3.5pt}
  \caption{Relative model performance on Tyson's data set during normal operating conditions; best value per metric is highlighted in boldface font}
  \label{tab:Tyson1}
  \begin{tabular}{|p{3.1em}|p{6.2em}|p{2.5em}p{2.5em}p{5em}|p{2.5em}p{2.5em}p{5em}|p{2.5em}p{2.5em}p{5em}|}
    \hline
    data set & Model 
    & \multicolumn{3}{c|}{15-min forecast}
    & \multicolumn{3}{c|}{30-min forecast}
    & \multicolumn{3}{c|}{60-min forecast} \\
    \cline{3-11}
      & & MAE& RMSE & MAPE  (\%) & MAE& RMSE & MAPE (\%) & MAE& RMSE & MAPE (\%)
    \\ 
    \hline
&STGCN & 2.54 & 4.68 & 5.84 & 2.76 & 5.03 & 7.10 &3.02 & 5.64 & 8.04\\
& GraphWaveNet & 2.42 & 4.56 & 5.61 & 2.74 & 5.01 & 6.98 & 2.98 & 5.42 & 8.51 \\
Tyson &ASTGCN  & 2.44 & 4.66 & 5.68 &  \textbf{2.52} & 4.62 & 6.06 & 2.70 & \textbf{4.95} & \textbf{6.69}\\
&DGCRN  & 2.38 & 4.47 & 5.43 & 2.55 & \textbf{4.58} & \textbf{6.01} &2.77 & 5.15& 7.35\\
&GCN-RWZ  & \textbf{2.25}  & \textbf{3.99} & \textbf{5.17} & 2.55 & 4.66 & 6.29 & \textbf{2.69} & 5.10 & 8.63  \\
    \hline
\end{tabular}
\end{table*}

\begin{table*}[!h]
    \renewcommand\arraystretch{1.1}
    \small
  \centering
  \setlength\tabcolsep{3.5pt}
  \caption{Relative model performance on Tyson's data set during work zone disruptions; best value per metric is highlighted in boldface font}
  \label{tab:Tyson2}
  \begin{tabular}{|p{3.1em}|p{6.2em}|p{2.5em}p{2.5em}p{5em}|p{2.5em}p{2.5em}p{5em}|p{2.5em}p{2.5em}p{5em}|}
    \hline
    data set & Model 
    & \multicolumn{3}{c|}{15-min forecast}
    & \multicolumn{3}{c|}{30-min forecast}
    & \multicolumn{3}{c|}{60-min forecast} \\
    \cline{3-11}
      & & MAE& RMSE & MAPE  (\%) & MAE& RMSE & MAPE (\%) & MAE& RMSE & MAPE (\%)
    \\ 
    \hline
&STGCN & 2.84 & 5.08 & 6.62 & 3.14 & 5.60 & 7.96 &3.57 & 6.46 & 9.52 \\

& GraphWaveNet & 2.74 & 5.11 & 6.59 & 2.96 & 5.21 & 7.42 & 3.37 & 5.90 & 8.02 \\
Tyson &ASTGCN  & 2.71 & 5.00 & 6.47 & 2.88 & 5.28 & 7.06 &\textbf{3.03} & \textbf{5.29} & \textbf{7.36} \\
&DGCRN  & 2.68 & 4.92& 6.51 & 2.90 & 5.33& 7.36& 3.25& 5.84& 7.65 \\
&GCN-RWZ  & \textbf{2.48}& \textbf{4.37}& \textbf{5.90}& \textbf{2.87} & \textbf{5.11} & \textbf{7.01} & 3.09 & 5.66 & 8.23 \\
    \hline
\end{tabular}
\end{table*}

The performance of all models deteriorates with increasing prediction length due to the nonlinear characteristics of traffic flow over long time intervals. The GCN-RWZ model shows excellent performance on both data sets, but particularly for the Richmond data set where it outperformed the other models for both normal and disrupted conditions across all forecast intervals. The benefits of the GCN-RWZ model are most apparent for longer term forecasts of 180 minutes. At this forecast length, the behavior of other GCN models is noticeably degraded under work zone disruptions. The accuracy of GCN-RWZ is degraded as well, but by a smaller magnitude. The results for the Tyson's data set are more mixed, but GCN-RWZ is competitive with baseline models across forecast lengths. 

Performance differences between the Richmond and Tyson's data set may be due to how missing data is imputed within those data sets. In the Tyson's data set, missing values are replaced with average historical speeds for a segment. And in the Tyson's data set there is a significant amount of missing data at night due to the fact that RITIS data is collected from vehicle probe systems. Night  is also when a significant amount of construction work occurs, in order to minimize traffic disruptions. This suggests that the impacts of some work zones are masked in the Tyson's data set.

 Again, the motivation for this work is to demonstrate how data fusion of heterogeneous graph feature maps can allow a GCN model to better learn complex traffic environments. The results on the Richmond data set, particularly Figure~\ref{fig:accuracy}, illustrate the benefit of this fusion.
 


\subsection{Detailed Evaluation of GCN-RWZ}
\begin{figure*}[htbp]
  \centering
  \begin{tabular}{cc}
  \includegraphics[width=3.5in]{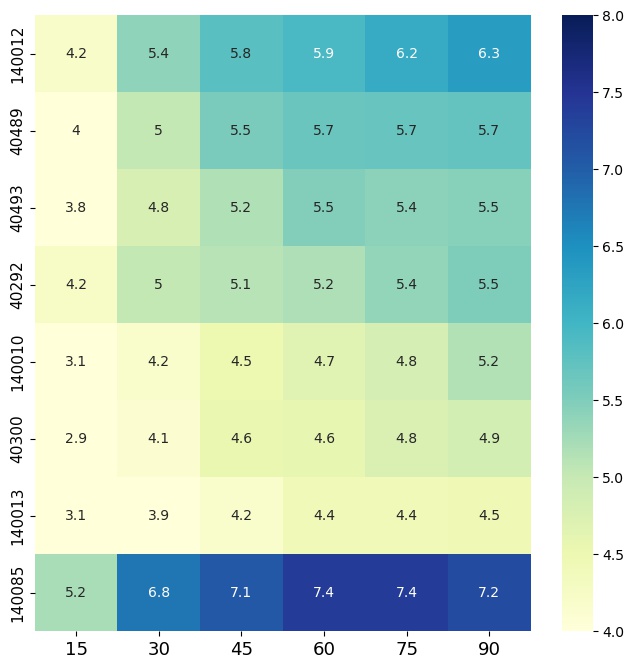} &    
  \includegraphics[width=3.5in]{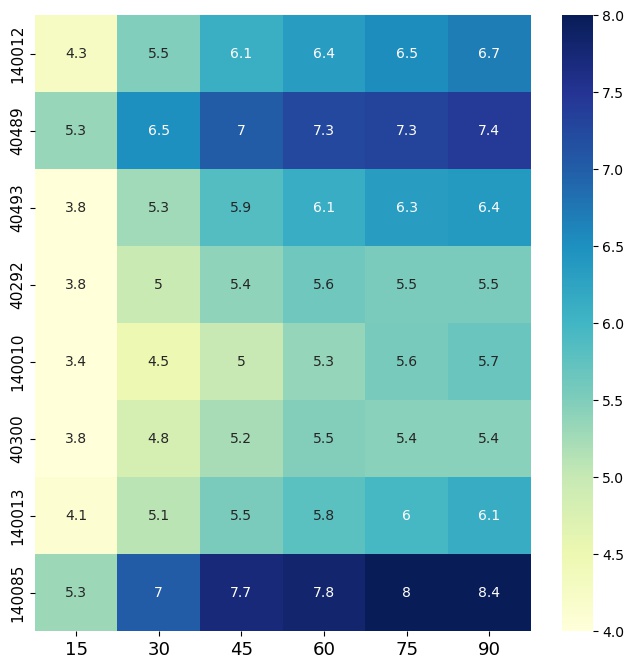}\\
  \end{tabular}
  \caption{RMSE heatmap representation of GCN-RWZ (left) and DGCRN (right) at forecast intervals (x-axis) over 8 random road segments (y-axis), Richmond data set under work zone disruption.}
  \label{heatmap}
\end{figure*}

We now provide a more detailed assessment of the predictive differences between GCN-RWZ and the best performing baseline model, DGCRN. In Figure~\ref{heatmap}, we randomly select 8 road segments from the Richmond data set and evaluate the average RMSE performance during workzone disruptions. Segments 140085, 140013, 140010, 40489, and 140012 are located in central Richmond and experience consistently high traffic volume throughout the day. Segments 40493, 40300, and 40292 are near an intersection in central Richmond. For short term prediction on some segments, the results do not differ much, though GCN-RWZ is consistently the better predictor and error is reduced by about 25\% ($\approx1$ MPH) for several segments. As the prediction length increases, the performance of both models worsens, as expected. However, the relative benefit of the workzone disruption information increases as the forecast interval increases. This is most notable for segment 40489, but it is apparent for most segments.  

\begin{figure}[!htbp]
\centering
\includegraphics[width=\columnwidth]{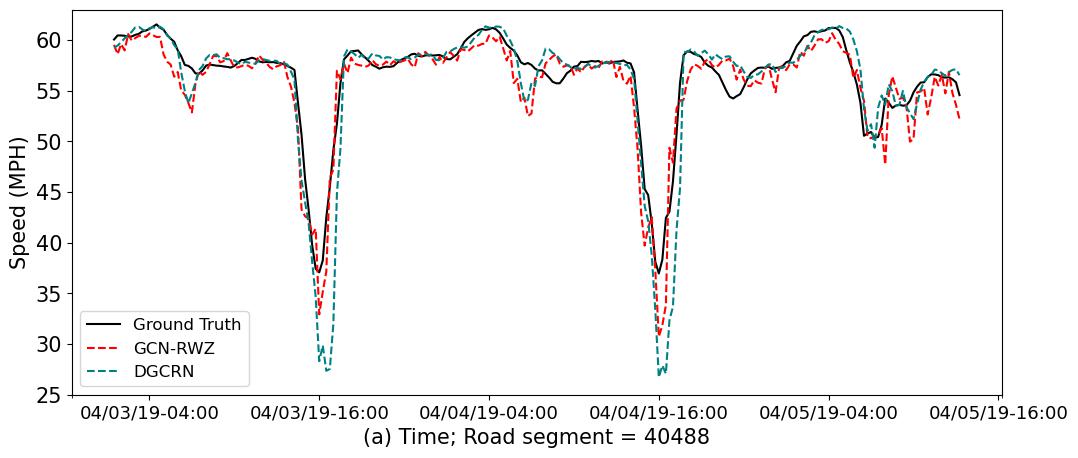}\\
\includegraphics[width=\columnwidth]{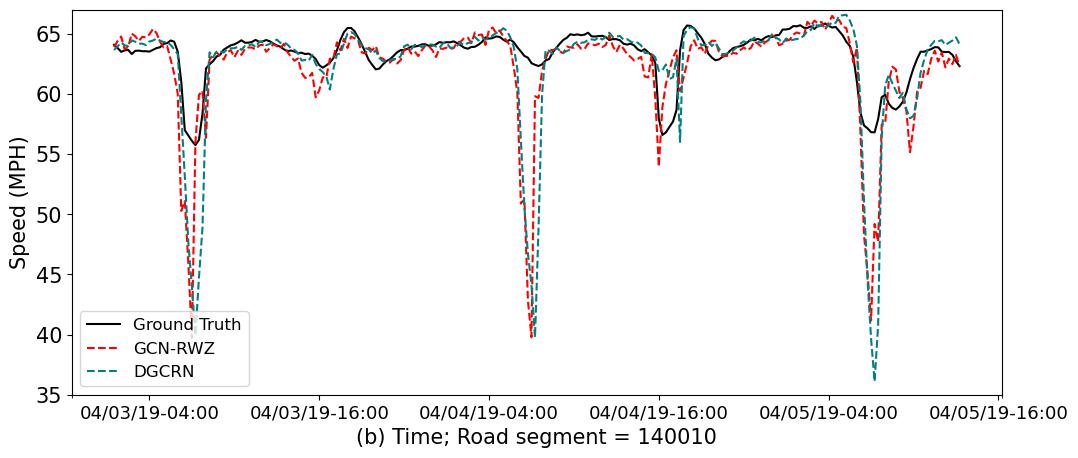}\\
\includegraphics[width=\columnwidth]{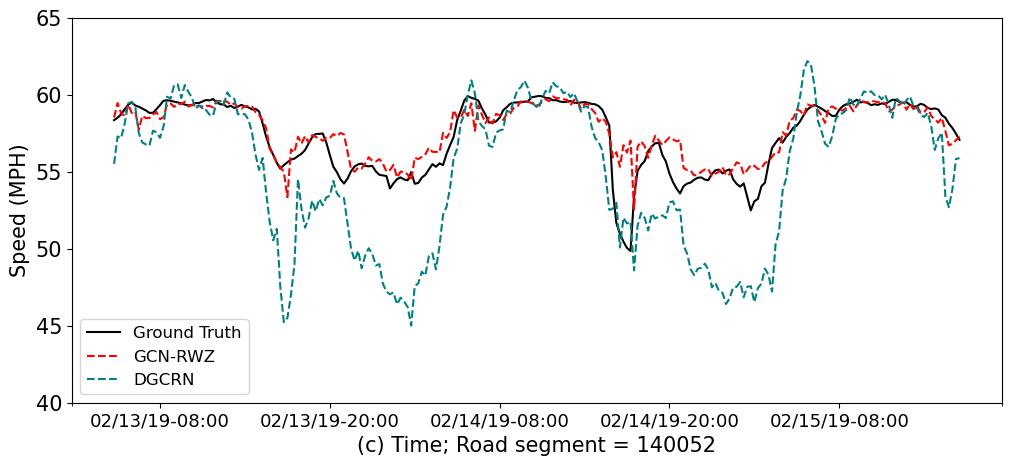}\\
\includegraphics[width=\columnwidth]{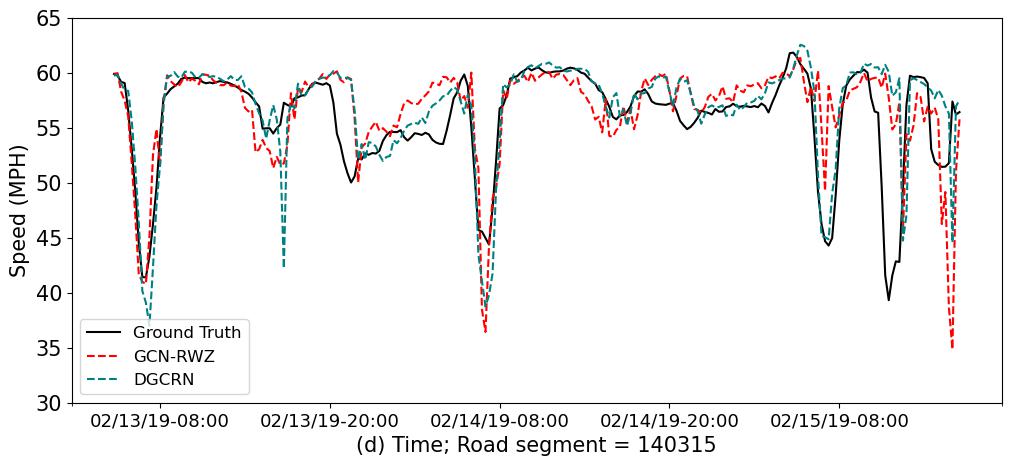}\\[-3mm]
\caption{Relative speed forecasting accuracy with 90 mins of forecast length on three road segments under the impact of the construction work from 04/03/2019 to 04/05/2019: (a) Road segment = "40488"; (b) Road segment = "140010" and  from 02/13/2019 to 02/15/2019: (c) Road segment = "140052";(d) Road segment = "140315".}
\label{predicttime}
\end{figure}

Figure~\ref{predicttime} shows how GCN-RWZ performs on forecasting lengths of 90 minutes during workzone disruptions, by superimposing the ground truth with the model-predicted speed on different road segments. The results for DGCRN are presented as well for comparison. For some disruptions, both models provide comparable results, though GCN-RWZ is generally a better predictor, particularly during rush hour times when speed decreases significantly. Both models overpredict a slowdown for segment 140010 during rush hour. The most notable difference between models is for segment 140052, where DGCRN performs far worse than GCN-RWZ at predicting the impacts of the workzone. In this case, the baseline model again overpredicts the magnitude of traffic slowdown on the segment.

\subsection{Ablation Study}
An ablation study was performed with regards to the data fusion mechanisms. The first ablation analysis considered the number of neighbors, H, in the hypergraph. As a comparison, we varied H as 1, 10, or all nodes as neighbors. As shown in~\ref{H}, the best performing configuration was for 5 nodes, though all results were relatively close. In our reported findings, the hypergraph is set to only observe the 5 nearest neighbors.

Another ablation analysis was undertaken to understand the sensitivity of the GCN-RWZ model to differences in speed wave formulations. Table~\ref{speedwave} shows the performance of the model using four different functions for feature fusion. The best RMSE, MAE, and MAPE are obtained on the first formulation of a learnable weight matrix for each feature map. This speed wave formulation is used throughout the reported results.

\begin{table}[!htbp]
  \renewcommand\arraystretch{1.3}
   \small
  \centering
  \setlength\tabcolsep{14pt}
\caption{Ablation study on values for different neighbors in hypergraphs; forecast length = 6 steps, normal traffic conditions}
\vspace{1mm}
\label{H}
 \begin{tabular}{p{4em}|p{2em}|p{2em}|p{5em}} 
 \hline
 Neighbors & MAE & RMSE & MAPE (\%) \\ [0.3ex]  \hline
H = 1 & 1.21  & 3.01  &  2.57  \\ 
 \hline
H = 5 & \textbf{1.16} & \textbf{2.92} & \textbf{2.46}  \\ 
 \hline
H = 10 & 1.18  & 2.95 &  2.52 \\
 \hline
H = ALL & 1.18  & 2.96 & 2.51  \\
 \hline
\end{tabular}
\end{table}

\begin{table}[htbp]
  \renewcommand\arraystretch{1.3}
   \small
  \centering
  \setlength\tabcolsep{9pt}
\caption{Ablation study on values for four speed wave $\hat{X_s}$ function; forecast length = 6 with WZ data}
\vspace{1mm}
\label{speedwave}
 \begin{tabular}{p{11em}|p{2em}|p{2em}|p{5em}} 
 \hline
 Speed Wave & MAE & RMSE & MAPE (\%) \\ [0.3ex]  \hline
$(W_s \odot X^s + W_c \odot X^c) \odot T^E$ & \textbf{1.25} & \textbf{3.13} & \textbf{2.78}  \\ 
 \hline
$W_s\odot X^s+W_c\odot X^c$ & 1.28 & 3.20 & 2.86  \\ 
 \hline
$X^s + W_c \odot X^c$& 1.33 & 3.29 & 2.97\\
 \hline
$X^s \odot X^s + W_c $& 1.35 & 3.32 & 2.99 \\
 \hline
\end{tabular}
\end{table}

\begin{table}[htbp]
    \renewcommand\arraystretch{1.3}
    \small
  \centering
  \setlength\tabcolsep{3pt}
  \caption{If it failed to find a solution, the distance is 3000 and duration is 2000}
  \label{tab:freq}
  \begin{tabular}{|p{2em}|p{2em}p{2em}|p{2em}p{2em}|p{2em}p{2em}|p{2em}p{2em}|}
    \hline
    & \multicolumn{2}{c|}{Scene 1}
    & \multicolumn{2}{c|}{Scene 2}
    & \multicolumn{2}{c|}{Scene 3}
    & \multicolumn{2}{c|}{Scene 4} \\
    \cline{2-9}
 & T & D &  T & D&  T & D &  T & D \\
    \hline
20 & 9.0 & 5.5 & 4.1 & 3.2  & x & x  & x & x \\
\hline
30 & 5.5 & x & x & x  & x & x  & x & x \\
\hline
40 & 4.1 & x & x & x  & x & x  & x & x \\
\hline
50 & 3.2 & x & x & x  & x & x  & x & x \\
\hline
\end{tabular}
\end{table} 
\section{Conclusions}
\label{sec:Conclusions}

The prediction of traffic flow under work zone disruptions is a topic of major interest for infrastructure managers. However state of the art GCN methods are not designed to account for workzones. We have proposed GCN-RWZ, a GCN-based model designed to integrate work zone disruption information into GCN models. In contrast to existing SOTA methods, the graph representation of traffic flow (speed) in GCN-RWZ is fused with a graph model of a construction workzone on any segment within the network. The fusion results in a time-history “speed wave” that serves as input to the GCN-RWZ learning algorithm. The GCN-RWZ fusion mechanism is flexible and generalizable, and could potentially be applied to other roadway information beyond work zone data. 
\\
The GCN-RWZ model was tested on two datasets designed for the explicit purpose of understanding work zone impacts on traffic flow. The performance is compared against baseline SOTA models that serve as established benchmarks for traffic flow prediction. The GCN-RWZ showed measurably better performance in traffic speed prediction compared to any of the benchmark models.  This suggests that the developed model is a viable platform for further studies, refinements, and implementation. However, the variations in performance observed for the Tyson's dataset suggest that how missing data is imputed, particularly during workzone events, may be impacting model performance, and highlights the need for additional open datasets. 

\IEEEpeerreviewmaketitle


\section*{Acknowledgment}
This work is supported in part from a grant to DL from the Virginia Transportation Research Council (VTRC). The authors would like to thank Michael Fitch and Michael Fontaine of VTRC for their guidance. This material is additionally based upon work by AS supported by (while serving at) the National Science Foundation. Any opinion, findings, and conclusions or recommendations expressed in this material are those of the author(s) and do not necessarily reflect the views of the National Science Foundation.

\section*{Data Availability}
The data used for this study was obtained through a license from the Virginia Transportation Research Council. The Richmond area dataset is available from: lrg.gmu.edu/GCNRWZ. The Tyson's Corner dataset is proprietary to the Virginia Department of Transportation and is only available by direct request. The source code developed through this project is available from: lrg.gmu.edu/GCNRWZ.

\ifCLASSOPTIONcaptionsoff
  \newpage
\fi

\bibliographystyle{IEEEtran}
\bibliography{ref.bib}

\begin{thebibliography}{10}
\providecommand{\url}[1]{#1}
\csname url@samestyle\endcsname
\providecommand{\newblock}{\relax}
\providecommand{\bibinfo}[2]{#2}
\providecommand{\BIBentrySTDinterwordspacing}{\spaceskip=0pt\relax}
\providecommand{\BIBentryALTinterwordstretchfactor}{4}
\providecommand{\BIBentryALTinterwordspacing}{\spaceskip=\fontdimen2\font plus
\BIBentryALTinterwordstretchfactor\fontdimen3\font minus \fontdimen4\font\relax}
\providecommand{\BIBforeignlanguage}[2]{{%
\expandafter\ifx\csname l@#1\endcsname\relax
\typeout{** WARNING: IEEEtran.bst: No hyphenation pattern has been}%
\typeout{** loaded for the language `#1'. Using the pattern for}%
\typeout{** the default language instead.}%
\else
\language=\csname l@#1\endcsname
\fi
#2}}
\providecommand{\BIBdecl}{\relax}
\BIBdecl

\bibitem{schrank_2019_2019}
D.~Schrank, B.~Eisele, and T.~Lomax, ``2019 urban mobility report,'' p.~50, 2019.

\bibitem{du2017predicting}
B.~Du, S.~Chien, J.~Lee, and L.~Spasovic, ``Predicting freeway work zone delays and costs with a hybrid machine-learning model,'' \emph{Journal of Advanced Transportation}, vol. 2017, 2017.

\bibitem{Zhang11}
``Data-driven intelligent transportation systems: A survey,'' vol.~12, no.~4, p. 1624–1639, 2011.

\bibitem{cui2019traffic}
Z.~Cui, K.~Henrickson, R.~Ke, and Y.~Wang, ``Traffic graph convolutional recurrent neural network: A deep learning framework for network-scale traffic learning and forecasting,'' \emph{IEEE Transactions on Intelligent Transportation Systems}, 2019.

\bibitem{li2017diffusion}
Y.~Li, R.~Yu, C.~Shahabi, and Y.~Liu, ``Diffusion convolutional recurrent neural network: Data-driven traffic forecasting,'' \emph{arXiv preprint arXiv:1707.01926}, 2017.

\bibitem{yu2017spatio}
B.~Yu, H.~Yin, and Z.~Zhu, ``Spatio-temporal graph convolutional networks: A deep learning framework for traffic forecasting,'' \emph{arXiv preprint arXiv:1709.04875}, 2017.

\bibitem{diao2019dynamic}
Z.~Diao, X.~Wang, D.~Zhang, Y.~Liu, K.~Xie, and S.~He, ``Dynamic spatial-temporal graph convolutional neural networks for traffic forecasting,'' in \emph{Proceedings of the AAAI Conference on Artificial Intelligence}, vol.~33, 2019, pp. 890--897.

\bibitem{guo2019attention}
S.~Guo, Y.~Lin, N.~Feng, C.~Song, and H.~Wan, ``Attention based spatial-temporal graph convolutional networks for traffic flow forecasting,'' in \emph{Proceedings of the AAAI Conference on Artificial Intelligence}, vol.~33, 2019, pp. 922--929.

\bibitem{tong2008highway}
M.~Tong and H.~Xue, ``Highway traffic volume forecasting based on seasonal arima model,'' \emph{Journal of Highway and Transportation Research and Development (English Edition)}, vol.~3, no.~2, pp. 109--112, 2008.

\bibitem{wu2004travel}
C.-H. Wu, J.-M. Ho, and D.-T. Lee, ``Travel-time prediction with support vector regression,'' \emph{IEEE transactions on intelligent transportation systems}, vol.~5, no.~4, pp. 276--281, 2004.

\bibitem{moretti2015urban}
F.~Moretti, S.~Pizzuti, S.~Panzieri, and M.~Annunziato, ``Urban traffic flow forecasting through statistical and neural network bagging ensemble hybrid modeling,'' \emph{Neurocomputing}, vol. 167, pp. 3--7, 2015.

\bibitem{kelejian1999generalized}
H.~H. Kelejian and I.~R. Prucha, ``A generalized moments estimator for the autoregressive parameter in a spatial model,'' \emph{International economic review}, vol.~40, no.~2, pp. 509--533, 1999.

\bibitem{huang2014deep}
W.~Huang, G.~Song, H.~Hong, and K.~Xie, ``Deep architecture for traffic flow prediction: deep belief networks with multitask learning,'' \emph{IEEE Transactions on Intelligent Transportation Systems}, vol.~15, no.~5, pp. 2191--2201, 2014.

\bibitem{tian2015predicting}
Y.~Tian and L.~Pan, ``Predicting short-term traffic flow by long short-term memory recurrent neural network,'' in \emph{2015 IEEE international conference on smart city/SocialCom/SustainCom (SmartCity)}.\hskip 1em plus 0.5em minus 0.4em\relax IEEE, 2015, pp. 153--158.

\bibitem{ma2015long}
X.~Ma, Z.~Tao, Y.~Wang, H.~Yu, and Y.~Wang, ``Long short-term memory neural network for traffic speed prediction using remote microwave sensor data,'' \emph{Transportation Research Part C: Emerging Technologies}, vol.~54, pp. 187--197, 2015.

\bibitem{ma2015large}
X.~Ma, H.~Yu, Y.~Wang, and Y.~Wang, ``Large-scale transportation network congestion evolution prediction using deep learning theory,'' \emph{PloS one}, vol.~10, no.~3, p. e0119044, 2015.

\bibitem{cui2018deep}
Z.~Cui, R.~Ke, Z.~Pu, and Y.~Wang, ``Deep bidirectional and unidirectional lstm recurrent neural network for network-wide traffic speed prediction,'' \emph{arXiv preprint arXiv:1801.02143}, 2018.

\bibitem{ma2017learning}
X.~Ma, Z.~Dai, Z.~He, J.~Ma, Y.~Wang, and Y.~Wang, ``Learning traffic as images: a deep convolutional neural network for large-scale transportation network speed prediction,'' \emph{Sensors}, vol.~17, no.~4, p. 818, 2017.

\bibitem{jo2018image}
D.~Jo, B.~Yu, H.~Jeon, and K.~Sohn, ``Image-to-image learning to predict traffic speeds by considering area-wide spatio-temporal dependencies,'' \emph{IEEE Transactions on Vehicular Technology}, vol.~68, no.~2, pp. 1188--1197, 2018.

\bibitem{vaswani2017attention}
A.~Vaswani, N.~Shazeer, N.~Parmar, J.~Uszkoreit, L.~Jones, A.~N. Gomez, {\L}.~Kaiser, and I.~Polosukhin, ``Attention is all you need,'' in \emph{Advances in neural information processing systems}, 2017, pp. 5998--6008.

\bibitem{wu2019graph}
Z.~Wu, S.~Pan, G.~Long, J.~Jiang, and C.~Zhang, ``Graph wavenet for deep spatial-temporal graph modeling,'' \emph{arXiv preprint arXiv:1906.00121}, 2019.

\bibitem{pan2019urban}
Z.~Pan, Y.~Liang, W.~Wang, Y.~Yu, Y.~Zheng, and J.~Zhang, ``Urban traffic prediction from spatio-temporal data using deep meta learning,'' in \emph{Proceedings of the 25th ACM SIGKDD International Conference on Knowledge Discovery \& Data Mining}, 2019, pp. 1720--1730.

\bibitem{ruiz2020gated}
L.~Ruiz, F.~Gama, and A.~Ribeiro, ``Gated graph recurrent neural networks,'' \emph{arXiv preprint arXiv:2002.01038}, 2020.

\bibitem{zheng2020gman}
C.~Zheng, X.~Fan, C.~Wang, and J.~Qi, ``Gman: A graph multi-attention network for traffic prediction,'' in \emph{Proceedings of the AAAI Conference on Artificial Intelligence}, vol.~34, no.~01, 2020, pp. 1234--1241.

\bibitem{keneshloo2019deep}
Y.~Keneshloo, T.~Shi, N.~Ramakrishnan, and C.~K. Reddy, ``Deep reinforcement learning for sequence-to-sequence models,'' \emph{IEEE transactions on neural networks and learning systems}, vol.~31, no.~7, pp. 2469--2489, 2019.

\bibitem{velivckovic2017graph}
P.~Veli{\v{c}}kovi{\'c}, G.~Cucurull, A.~Casanova, A.~Romero, P.~Lio, and Y.~Bengio, ``Graph attention networks,'' \emph{arXiv preprint arXiv:1710.10903}, 2017.

\bibitem{bacciu2020gentle}
D.~Bacciu, F.~Errica, A.~Micheli, and M.~Podda, ``A gentle introduction to deep learning for graphs,'' \emph{Neural Networks}, vol. 129, pp. 203--221, 2020.

\bibitem{perozzi2014deepwalk}
B.~Perozzi, R.~Al-Rfou, and S.~Skiena, ``Deepwalk: Online learning of social representations,'' in \emph{Proceedings of the 20th ACM SIGKDD international conference on Knowledge discovery and data mining}, 2014, pp. 701--710.

\bibitem{kang2019learning}
Z.~Kang, H.~Xu, J.~Hu, and X.~Pei, ``Learning dynamic graph embedding for traffic flow forecasting: A graph self-attentive method,'' in \emph{2019 IEEE Intelligent Transportation Systems Conference (ITSC)}.\hskip 1em plus 0.5em minus 0.4em\relax IEEE, 2019, pp. 2570--2576.

\bibitem{wu2020comprehensive}
Z.~Wu, S.~Pan, F.~Chen, G.~Long, C.~Zhang, and S.~Y. Philip, ``A comprehensive survey on graph neural networks,'' \emph{IEEE Transactions on Neural Networks and Learning Systems}, 2020.

\bibitem{zhao2019t}
L.~Zhao, Y.~Song, C.~Zhang, Y.~Liu, P.~Wang, T.~Lin, M.~Deng, and H.~Li, ``T-gcn: A temporal graph convolutional network for traffic prediction,'' \emph{IEEE Transactions on Intelligent Transportation Systems}, 2019.

\bibitem{zhou2020reinforced}
F.~Zhou, Q.~Yang, K.~Zhang, G.~Trajcevski, T.~Zhong, and A.~Khokhar, ``Reinforced spatio-temporal attentive graph neural networks for traffic forecasting,'' \emph{IEEE Internet of Things Journal}, 2020.

\bibitem{oneto2022towards}
L.~Oneto, N.~Navarin, B.~Biggio, F.~Errica, A.~Micheli, F.~Scarselli, M.~Bianchini, L.~Demetrio, P.~Bongini, A.~Tacchella \emph{et~al.}, ``Towards learning trustworthily, automatically, and with guarantees on graphs: An overview,'' \emph{Neurocomputing}, 2022.

\bibitem{du2020traffic}
B.~Du, X.~Hu, L.~Sun, J.~Liu, Y.~Qiao, and W.~Lv, ``Traffic demand prediction based on dynamic transition convolutional neural network,'' \emph{IEEE Transactions on Intelligent Transportation Systems}, vol.~22, no.~2, pp. 1237--1247, 2020.

\bibitem{zhu2022kst}
J.~Zhu, X.~Han, H.~Deng, C.~Tao, L.~Zhao, P.~Wang, T.~Lin, and H.~Li, ``Kst-gcn: A knowledge-driven spatial-temporal graph convolutional network for traffic forecasting,'' \emph{IEEE Transactions on Intelligent Transportation Systems}, 2022.

\bibitem{luo2022evaluating}
Y.-K. Luo, S.-X. Chen, L.~Zhou, and Y.-Q. Ni, ``Evaluating railway noise sources using distributed microphone array and graph neural networks,'' \emph{Transportation Research Part D: Transport and Environment}, vol. 107, p. 103315, 2022.

\bibitem{li2022graph}
H.~Li, J.~Zhang, L.~Yang, J.~Qi, and Z.~Gao, ``Graph-gan: A spatial-temporal neural network for short-term passenger flow prediction in urban rail transit systems,'' \emph{arXiv preprint arXiv:2202.06727}, 2022.

\bibitem{wang2019heterogeneous}
X.~Wang, H.~Ji, C.~Shi, B.~Wang, Y.~Ye, P.~Cui, and P.~S. Yu, ``Heterogeneous graph attention network,'' in \emph{The world wide web conference}, 2019, pp. 2022--2032.

\bibitem{defferrard2016convolutional}
M.~Defferrard, X.~Bresson, and P.~Vandergheynst, ``Convolutional neural networks on graphs with fast localized spectral filtering,'' in \emph{Advances in neural information processing systems}, 2016, pp. 3844--3852.

\bibitem{hammond2011wavelets}
D.~K. Hammond, P.~Vandergheynst, and R.~Gribonval, ``Wavelets on graphs via spectral graph theory,'' \emph{Applied and Computational Harmonic Analysis}, vol.~30, no.~2, pp. 129--150, 2011.

\bibitem{kipf2016semi}
T.~N. Kipf and M.~Welling, ``Semi-supervised classification with graph convolutional networks,'' \emph{arXiv preprint arXiv:1609.02907}, 2016.

\bibitem{li2021dynamic}
F.~Li, J.~Feng, H.~Yan, G.~Jin, F.~Yang, F.~Sun, D.~Jin, and Y.~Li, ``Dynamic graph convolutional recurrent network for traffic prediction: Benchmark and solution,'' \emph{ACM Transactions on Knowledge Discovery from Data (TKDD)}, 2021.

\end{thebibliography}
\end{document}